\documentclass[11pt,letterpaper]{article}
\usepackage{ijcnlp2017}
\usepackage{times}
\usepackage{latexsym}
\usepackage{color, colortbl}
\usepackage{url}

\definecolor{Gray}{gray}{0.95}
\usepackage{amsmath}
\usepackage{graphicx}
\usepackage{array,multirow,graphicx}
\usepackage{color}

\ijcnlpfinalcopy

\def\ijcnlppaperid{197}

\newcommand{\cev}[1]{\reflectbox{\ensuremath{\vec{\reflectbox{\ensuremath{#1}}}}}}

\title{ Multilingual Hierarchical Attention Networks for Document Classification }

\author{Nikolaos Pappas \\
   Idiap Research Institute \\
	Rue Marconi 19\\
   CH-1920 Martigny, Switzerland\\
  {\tt \small nikolaos.pappas@idiap.ch} \\\And
  Andrei Popescu-Belis\\
  HEIG-VD / HES-SO \\
	Route de Cheseaux 1\\
  CH-1401 Yverdon, Switzerland\\
  {\tt \small andrei.popescu-belis@heig-vd.ch} \\}

\date{}

\begin{document}
\maketitle
\begin{abstract}
Hierarchical attention networks have recently achieved remarkable performance for document classification in a given language.  However, when multilingual document collections are considered, training such models separately for each language entails linear parameter growth and lack of cross-language transfer. Learning a single multilingual model with fewer parameters is therefore a challenging but potentially beneficial objective. To this end, we propose multilingual hierarchical attention networks for learning document structures, with shared encoders and/or shared attention mechanisms across languages, using multi-task learning and an aligned semantic space as input.  We evaluate the proposed models on multilingual document classification with disjoint label sets, on a large dataset which we provide, with 600k news documents in 8 languages, and 5k labels.  The multilingual models outperform monolingual ones in low-resource as well as full-resource settings, and use fewer parameters, thus confirming their computational efficiency and the utility of cross-language transfer.
\end{abstract}

\section{Introduction}

Learning word sequence representations has become increasingly useful for a variety of NLP tasks such as document classification \citep{tang15,yang16}, neural machine translation (NMT) \citep{cho14,thang15}, question answering \citep{Sukhbaatar15,kumar15} and summarization \citep{rush15}.
However, when data are available in multiple languages, representation learning must address two main challenges. Firstly, the computational cost of training separate models for each language, which grows linearly with their number, or even quadratically in the case of multi-way multilingual NMT \citep{firat16}. Secondly, the models should be capable of cross-language transfer, which is an important component in human language learning \citep{ringbom2007}. For instance,  \citet{Johnson16} attempted to use a single sequence-to-sequence neural network model for NMT across multiple language pairs.

\begin{figure}[t]
\centering
\includegraphics[scale=0.15]{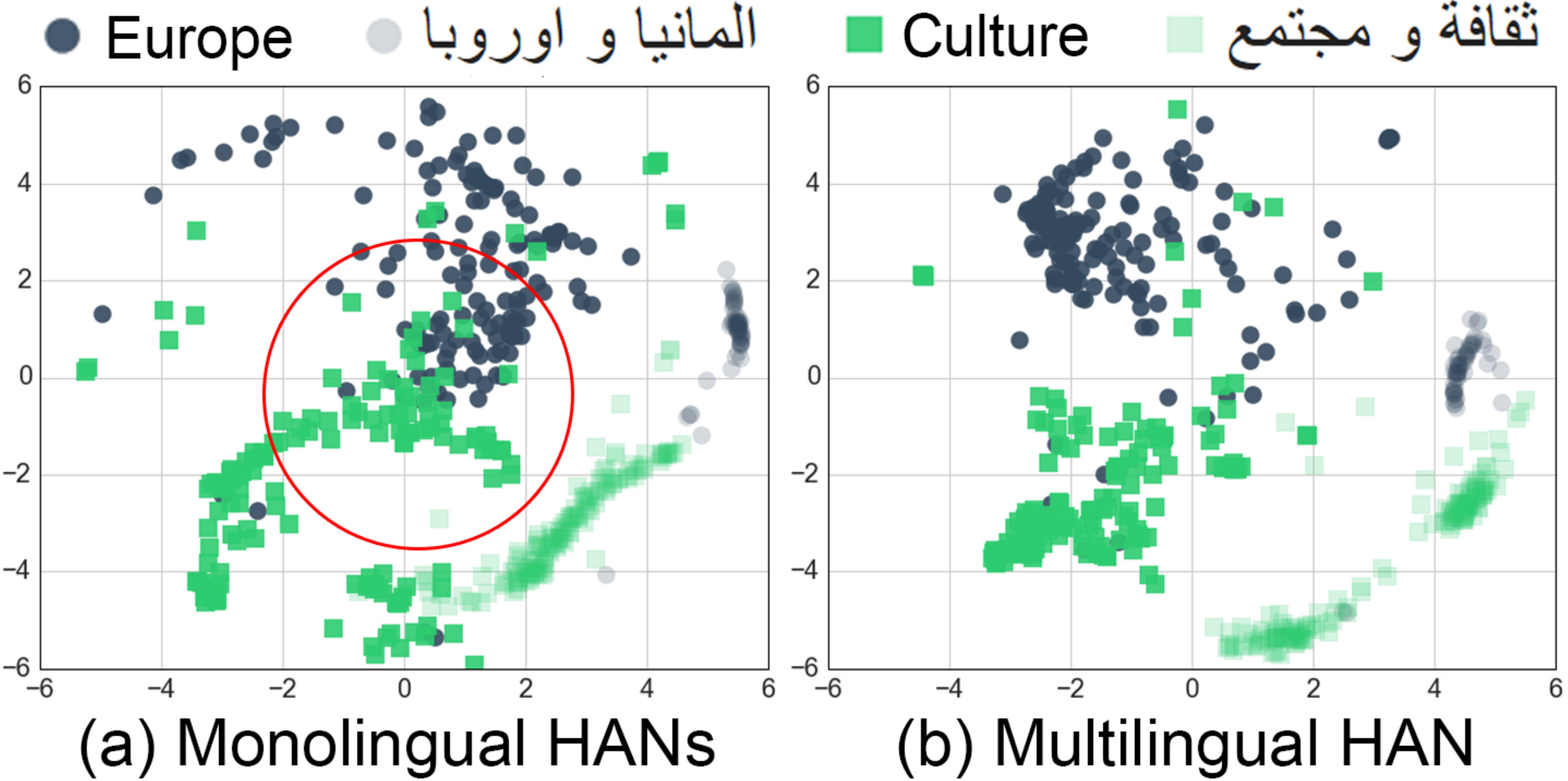}
\caption{Vectors of documents labeled with \textit{`Europe'}, \textit{`Culture'} and their Arabic counterparts.  The multilingual hierarchical attention network separates topics better than monolingual ones.}
\label{tsne_projection}
\end{figure}

Previous studies in document classification attempted to address these issues by employing multilingual word embeddings, which allow direct comparisons and groupings across languages \citep{klementiev12,hermann14,ferreira16}. However, they are only applicable when common label sets are available across languages  which is often not the case (e.g.\ Wikipedia or news). Moreover, despite recent advances in monolingual document modeling \citep{tang15,yang16}, multilingual models are still based on shallow approaches.

In this paper, we propose \textit{Multilingual Hierarchical Attention  Networks} to learn shared document structures across languages for document classification with disjoint label sets, as opposed to training language-specific training of hierarchical attention networks (HANs) \citep{yang16}. Our networks have a hierarchical structure with word and sentence encoders, along with attention mechanisms. Each of these can either be shared across languages or kept language-specific. To enable cross-language transfer, the networks are trained with multi-task learning across languages using an aligned semantic space as input. Fig.~\ref{tsne_projection} displays document vectors, projected with t-SNE \citep{tsne}, for two topics and two languages, either learned by monolingual HANs~(a) or by our multilingual HAN~(b).  The multilingual HAN achieves better separation between `Europe' and `Culture' topics in English as a result of the knowledge transfer from Arabic.

We evaluate our model against strong monolingual baselines, in low-resource and full-resource scenarios, on a large multilingual document collection with 600k documents, labeled with general (1.2k) and specific topics (4.4k), in 8 languages from Deutsche Welle's news website.\footnote{Germany's news broadcaster:  \url{http://dw.com}.} Our multilingual models outperform monolingual ones in both scenarios, thus confirming the utility of cross-language transfer and the computational efficiency of the proposed architecture. To encourage further research in multilingual representation learning our code and dataset are made available at \url{https://github.com/idiap/mhan}.

\section{Related Work}
\label{sec:related-work}

Research on \textit{learning multilingual word representations} is based on early work on word embeddings \cite{turian10,word2vec,pennington14}. The goal is to learn an aligned word embedding space for multiple languages by leveraging bilingual dictionaries \cite{faruqui14,ammar16}, parallel sentences \cite{gows15} or comparable documents such as Wikipedia pages \citep{yih11,rfou13}.  Bilingual embeddings were learned from word alignments using neural language models \citep{klementiev12,zou13}, including auto-encoders \citep{chandar14}.  Despite progress at the word level, the document level remains comparatively less explored.  The approaches proposed by \citet{hermann14} or \citet{ferreira16} are based on shallow modeling and are applicable only to classification tasks with label sets shared across languages, which are costly to produce and are often unavailable. Here, we remove this constraint, and develop deeper multilingual document models with hierarchical structure based on prior art at the word level.

Early work on \textit{neural document classification} was based on shallow feed-forward networks, which required unsupervised pre-training \citep{le14}. Later studies focused on neural networks with hierarchical structure. \citet{kim14} proposed a convolutional neural network (CNN) for sentence classification. \citet{johnson14} proposed a CNN for high-dimensional data classification, while \citet{zhang15} adopted a character-level CNN for text classification. \citet{lai15} proposed a recurrent CNN to capture sequential information, which outperformed simpler CNNs. \citet{lin15} and \citet{tang15} proposed hierarchical recurrent NNs and showed that they were superior to CNN-based models. Recently, \citet{yang16} proposed a hierarchical attention network (HAN) with bi-directional gated encoders which outperforms traditional and neural baselines. Using such networks in multilingual settings has two drawbacks: the computational complexity increases linearly with the number of languages, and knowledge is acquired separately for each language. We address these issues by proposing a new multilingual model based on HANs, which learns shared document structures and to transfer knowledge across languages.

Early examples of \textit{attention mechanisms} appeared in computer vision, e.g.\ for optical character recognition \citep{Larochelle10}, image tracking \citep{denil12}, or image classification \cite{mnih14}. For text classification, studies which aimed to learn the importance of sentences included those by \citet{yessenalina10,emnlp14,yang16} and more recently those by \citet{Pappas2017b,ji17}.  For NMT, \citet{Bahdanau15} proposed an attention-based encoder-decoder network, while \citet{thang15} proposed a local and ensemble attention model. \citet{firat16} proposed a single encoder-decoder model with shared attention across language pairs for multi-way, multilingual NMT. \citet{herman15} developed attention-based document readers for question answering.  \citet{Sukhbaatar15} proposed a recurrent attention model over an external memory. Similarly, \citet{kumar15} introduced a dynamic memory network for question answering and other tasks. We propose here to share attention across languages, at one or more levels of hierarchical document models, which, to our knowledge, has not been attempted before.

\section{Background: Hierarchical Attention Networks for Document Classification}
\label{sec:background}

\begin{figure}[ht]
\centering
\hspace{-4mm}\includegraphics[scale=0.62]{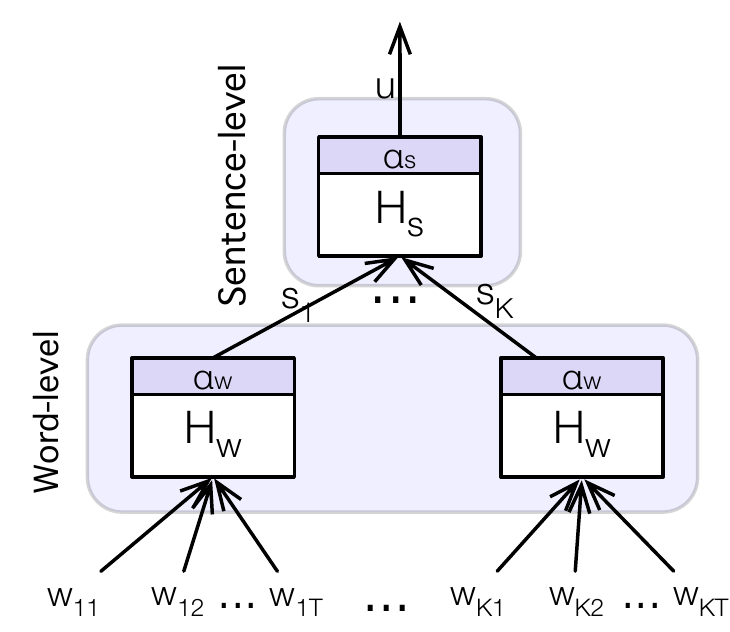}
\caption{General architecture of hierarchical attention neural networks for modeling documents.}
\label{general_architecture}
\end{figure}

We adopt a general hierarchical attention architecture for document representation, displayed in Figure~\ref{general_architecture}, which is derived from the one proposed by \citet{yang16}.  Our architecture is general in the sense that it defines only the hierarchical structure, but accommodates different types of individual components, i.e.\ encoders and attention models. 
We consider a dataset $D=\{(x_i, y_i), i=1, \ldots, N\}$ made of $N$ documents $x_i$ with labels $y_i \in \{0,1\}^k$. Each document is represented by the sequence of $d$-dimensional embeddings of their words grouped into sentences, $x_{i}=\{w_{11},w_{12}, \ldots, w_{KT}\}$, $T$ being the maximum number of words in a sentence, and $K$ the maximum number of sentences in a document.

The network takes as input a document $x_i$ and outputs a document vector $u_i$. In particular, it has two levels of abstraction, word vs.\ sentence. The word level is made of an encoder $g_{w}$ with parameters $H_w$ and an attention model $a_{w}$ with parameters $A_w$, while the sentence level similarly includes an encoder and an attention model ($g_{s}$, $H_s$ and $a_s$, $A_s$). The output $u_i$ is used by the classification layer to determine $y_i$.

\subsection{Encoder Layers}
At the word level, the function $g_w$ encodes the sequence of input words $\{w_{it}\ |\ t=1, \ldots, KT\}$ for each sentence $i$ of the document, noted as:
\begin{gather}
  h^{(it)}_{w} = \{g_{w}(w_{it}) |\ t=1, \ldots, K\}
\end{gather}
\noindent At the sentence level, after combining the intermediate word vectors  $\{h^{(it)}_{w}\ |\  t=1, \ldots, T\}$ to a sentence vector $s_{i}$ (as explained in~\ref{subatt}),  the function $g_s$ encodes the sequence of
sentence vectors $\{s_{i}\ |\ i=1, \ldots, K\}$, noted as $h^{(i)}_{s}$.

The $g_w$ and $g_s$ functions can be any feed-forward or recurrent networks with parameters $H_w$ and $H_s$ respectively.
We consider the following networks: a fully-connected one, noted as \textsc{Dense}, 
a Gated Recurrent Unit network \cite{cho14} noted as GRU\footnote{GRU is a simplified version of Long-Short Term Memory, LSTM \cite{hoch97}.}, and a bi-directional GRU which captures temporal information forward or backward in time, noted as biGRU. The latter is defined as a concatenation of the hidden states for each input vector obtained from the forward GRU, $\vec{g_w}$, and the backward GRU, $\cev{g_w}$:
  \begin{equation}
    h^{(it)}_{w}  = \big[ \vec{g_w}(h^{(it)}_{w});  \cev{g_w}(h^{(it)}_{w}) \big].
  \end{equation}
  \noindent The same concatenation is applied for the hidden-state representation of a sentence $h^{(i)}_{s}$.

\subsection{Attention Layers}
\label{subatt}

A typical way to obtain a representation for a given word sequence at each level is by taking the last hidden-state vector that is output by the encoder. However, it is hard to encode all the relevant input information needed in a   fixed-length vector.
This problem is addressed by introducing an attention mechanism at each level (noted $\alpha_{w}$ and $\alpha_{s}$) that estimates the importance of each hidden state vector to the representation of the sentence or document meaning respectively. The sentence vector $s_{i} \in R^{d_{w}}$, where $d_{w}$ is the  dimension of the word encoder, is thus obtained as follows:
\begin{equation}
\frac{1}{T} \sum^T_{t=1} \alpha^{(it)}_{w} h^{(it)}_{w} = \frac{1}{T} \sum^T_{t=1} \frac{\mathrm{exp}(v^\top_{it} u_w)}{\sum_{j}\mathrm{exp}(v^\top_{ij} u_w)}  h^{(it)}_{w}
\label{att_eq}
\end{equation}
\noindent where $v_{it} = f_{w}(h^{(it)}_{w})$ is a fully-connected neural network with $W_w$ parameters.
Similarly, the document vector $u \in R^{d_{s}}$, where $d_{s}$ is the dimension of the sentence encoder, is obtained as follows:
\begin{equation}
\frac{1}{K} \sum^{K}_{i=1} \alpha^{(i)}_{s}  h^{(i)}_{s} = \frac{1}{K} \sum^K_{i=1} \frac{\mathrm{exp}(v^\top_{i} u_s)}{\sum_{j}\mathrm{exp}(v^\top_{j} u_s)} h^{(i)}_{s}
\label{att_eq2}
\end{equation}
\noindent where $v_{i} = f_{s}(h^{(i)}_{s})$ is a fully-connected neural network with $W_s$ parameters. The vectors $u_w$ and $u_s$ are parameters which encode the word context and sentence context respectively, and are learned jointly with the rest of the parameters. The total set of parameters for $a_{w}$ is $A_w = \{ W_{w}, u_{w}\}$ and for $a_{s}$ is $A_s = \{W_{s}, u_{s}\}$.

\subsection{Classification Layers}

The output of such a network is typically fed to a softmax layer for classification, with a loss based on the cross-entropy between gold and predicted labels \cite{tang15} or on the negative log-likelihood of the correct labels \cite{yang16}. However,  softmax overemphasizes the probability of the most likely label, which may not be ideal for multi-label classification because each document should have more than one likely labels independent of each other, as we verified empirically in our preliminary experiments.  Hence, we replace the softmax with a sigmoid function, so that for each document $i$ represented by the vector $u_i$ we model the probability of the $k$ labels as follows:
\begin{equation}\label{eq:classification}
  \hat{y_{i}} = p(y|u_i) = \frac{1}{1+e^{-(W_{c} u_i + b_{c}) }}  \in [0,1]^k
\end{equation}
\noindent where $W_{c}$ is a $d_{s} \times k$ matrix and $b_{c}$ is the bias term for the classification layer. The training loss based on cross-entropy is computed as follows:\vspace{-0.8em}
\begin{equation}
\mathcal{L}(\theta) = - \frac{1}{N} \sum^N_{i=1} \mathcal{H}(y_{i}, \hat{y}_{i})
\label{eq:mono-objective}
\end{equation}
\noindent where $\theta$
is a notation for all the parameters of the model (i.e.\ $H_w, A_w, H_s, A_s, W_c$), and $\mathcal{H}$ is the binary cross-entropy of the gold labels $y_{i}$ and predicted labels $\hat{y}_{i}$ for a  document $i$. The above objective is differentiable and can be minimized with stochastic gradient descent (SGD) \citep{SGD} or variants such as Adam \cite{adam}, to maximize classification performance.

\section{Multilingual Hierarchical Attention Networks: MHANs}
\label{sec:proposed-model}

When multilingual data is available, the above network can be trained on each language separately, but in this case  the needed parameters grow linearly with the number of languages. Moreover, this does not exploit common knowledge across languages or to transfer it from one to another. We propose here a HAN with shared components across languages, which has slower parameter growth (hence sublinear) compared to monolingual ones and enables knowledge transfer across languages.  We now consider $M$ languages noted $L=\{L_{l}\ |\ l=1, \ldots, M\}$, and a multilingual set of topic-labeled documents $D_{l}=\{(x^{(l)}_i, y^{(l)}_i)\ |\ i=1, \ldots, N_{l}, l=1,...,M\}$  defined as above (Section~\ref{sec:background}).

\subsection{Sharing Components across Languages}
\label{sec:sharing-components}

\begin{figure}[t]
\centering
\hspace{0mm}\includegraphics[scale=0.57]{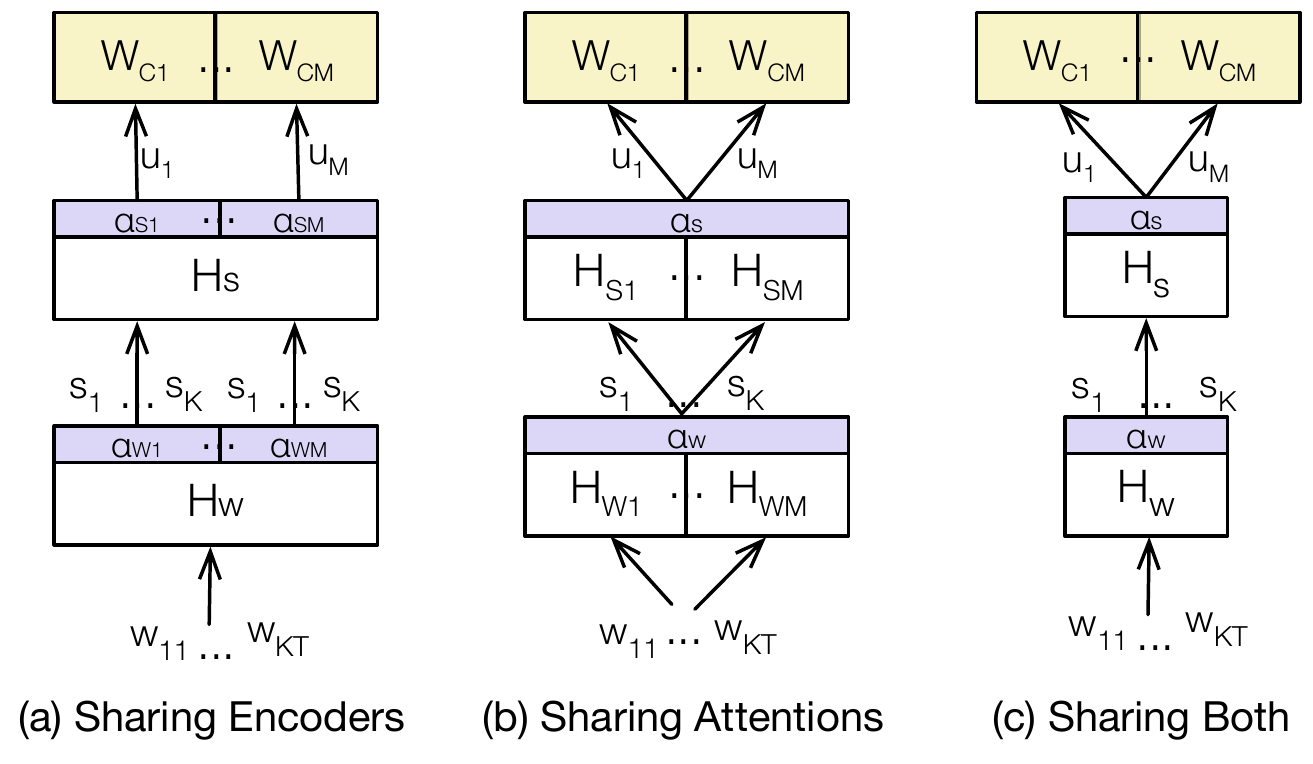}
\caption{Multilingual hierarchical attention networks for modeling documents and classifying them over disjoint label sets.}
\label{multilingual_architecture}
\end{figure}

To enable multilingual learning, we propose three distinct ways for sharing components between networks in a multi-task learning setting, depicted in Figure~\ref{multilingual_architecture}, namely:
(a)~sharing the parameters of word and sentence encoders, noted $\theta_{enc}=\{H_w, W^{(l)}_w, H_s, W^{(l)}_s, W^{(l)}_c\}$; (b)~sharing the parameters of word and sentence attention models, noted $\theta_{att}=\{H^{(l)}_w, W_w, H^{(l)}_s, W_s, W^{(l)}_c\}$; and (c)~sharing both previous sets of parameters, noted $\theta_{both}=\{H_w, W_w, H_s, W_s, W^{(l)}_c\}$. For instance, the document representation of a text for language $l$ based on a shared sentence-level attention would be computed based on Eq.~\ref{att_eq2} by using the same parameters $W_{s}$ and $u_s$ across languages.

Let $\theta_{mono}=\{H_w^{(l)}, W_w^{(l)}, H_s^{(l)}, W_s^{(l)}, W^{(l)}_c\}$ be the parameters of multiple independent monolingual models
with \textsc{Dense} encoders, then we have:
\vspace{-1em}
\begin{equation}
  |\theta_{mono}| >  |\theta_{enc}| > |\theta_{att}| > |\theta_{both}|
\end{equation}
\noindent where $|\cdot|$ is the number of parameters in a set.
For GRU and biGRU encoders, the inequalities still hold, but swapping $|\theta_{enc}|$ and $|\theta_{att}|$. Excluding the classification layer which is
necessarily language-specific, the (a) and (b) networks have sublinear numbers of parameters and the (c) network has a constant number of parameters with respect to the number of languages.\begin{table}[htp]
	\small
  \centering
\begin{tabular}{ c || c |c | c || c |  c }
 Languages &  \multicolumn{3}{c||}{Documents}  & \multicolumn{2}{c}{Labels}\\ \hline
        $L$ & $|X|$ & $\bar{s}$  & $\bar{w}$ &  $|Y_g|$ &  $|Y_s|$ \\ \hline
    English & 112,816 & 17.9 & 516.2 & 327 & 1,058 \\
    German  & 132,709 & 22.3 & 424.1 & 367 & 809 \\
    Spanish & 75,827 & 13.8 & 412.9 & 159 & 684 \\
    Portuguese & 39,474 & 20.2 & 571.9  & 95 & 301 \\
    Ukrainian & 35,423 & 17.6 & 342.9 & 28 & 260 \\
    Russian & 108,076 & 16.4 & 330.1 & 102 & 814 \\
    Arabic  & 57,697  & 13.3 & 357.7 & 91 & 344 \\
    Persian & 36,282 & 18.7 & 538.4 & 71 & 127 \\ \hline
		All & 598,304	& 17.52 & 436.7 & 1,240	& 4,397 \\ \hline
\end{tabular}
\caption{\label{dw_stats}Statistics of the Deutsche Welle corpus: $\bar{s}$ and $\bar{w}$ are the average numbers of sentences and words per document.}
\end{table} The word embeddings are not considered as parameters in our setup because they are fixed during training. For learned word embeddings, the argument still holds if we consider their parameters as part of the word-level encoder.

Depending on the label sets, several types of document classification problems can be solved with such architectures.  First, label sets can be common or disjoint across languages. Second, considering labels as $k$-hot vectors, $k=1$ corresponds to a multi-class task, while $k>1$ is a multi-label task.  We focus here on the multi-label problem with disjoint label sets. Moreover, we assume an aligned input space i.e.\ with multilingual word embeddings that have aligned meanings across languages \citep{ammar16}. With non-aligned word embeddings, the multilingual transfer is harder due to the lack of parallel information, as we show in Section~\ref{sec:results}, Table~\ref{multi8}.

\begin{table*}
  \setlength{\tabcolsep}{4.3pt}
	\small
  \centering
  \begin{tabular}{ p{0.4cm} | p{0.2cm} l | c c c c c c c | c c c c c c c}
  & & &  \multicolumn{7}{c|}{English + Auxiliary $\rightarrow$ English} &  \multicolumn{7}{c}{English + Auxiliary $ \rightarrow$ Auxiliary} \\ \hline
     &  & Models & de  & es  & pt & uk & ru & ar & fa & de & es & pt & uk & ru & ar & fa  \\ \hline
      \parbox[t]{1mm}{\multirow{6}{*}{\rotatebox[origin=c]{90}{  $Y_\mathit{general}$  }}}
      &  \parbox[t]{1mm}{\multirow{3}{*}{\rotatebox[origin=c]{90}{  Mono  }}} & NN (Avg)  &  \multicolumn{7}{c|}{ \rule[0.25em]{0.3\columnwidth}{0.3pt} 50.7 \rule[0.25em]{0.3\columnwidth}{0.3pt} } & 53.1 & 70.0 & 57.2 & 80.9 & 59.3 & 64.4 & 66.6 \\
      & & HNN (Avg) &  \multicolumn{7}{c|}{ \rule[0.25em]{0.3\columnwidth}{0.3pt} 70.0 \rule[0.25em]{0.3\columnwidth}{0.3pt} } & 67.9 & 82.5 & 70.5 & 86.8 & 77.4 & 79.0 & 76.6 \\
      &  & HAN (Att)  & \multicolumn{7}{c|}{ \rule[0.25em]{0.3\columnwidth}{0.3pt} 71.2 \rule[0.25em]{0.3\columnwidth}{0.3pt} } & 71.8 & 82.8 & 71.3 & 85.3 & 79.8 & 80.5 & 76.6 \\ \cline{2-17}
       & \parbox[t]{2mm}{\multirow{3}{*}{\rotatebox[origin=c]{90}{  Multi  }}} &  MHAN-Enc  & 71.0 & 69.9 & 69.2 & 70.8 & 71.5 & 70.0 & 71.3 & 69.7 & \textbf{82.9} & 69.7 & 86.8 & 80.3 & 79.0 & 76.0  \\
    & &  MHAN-Att &  \textbf{74.0} & \textbf{74.2} & \textbf{74.1} & \textbf{72.9} & \textbf{73.9} & \textbf{73.8} & \textbf{73.3} & \textbf{72.5} & 82.5 & 70.8 & \textbf{87.7} & 80.5 & \textbf{82.1} & 76.3 \\
    & &  MHAN-Both &  72.8 & 71.2 & 70.5 & 65.6 & 71.1 & 68.9 & 69.2 & 70.4 & 82.8 & \textbf{71.6} & 87.5 & \textbf{80.8} & 79.1 & \textbf{77.1} \\ \hline
    \parbox[t]{1mm}{\multirow{6}{*}{\rotatebox[origin=c]{90}{  $Y_\mathit{specific}$  }}} &  \parbox[t]{2mm}{\multirow{3}{*}{\rotatebox[origin=c]{90}{  Mono  }}}  &  NN (Avg)  & \multicolumn{7}{c|}{ \rule[0.25em]{0.3\columnwidth}{0.3pt} 24.4 \rule[0.25em]{0.3\columnwidth}{0.3pt} }  & 21.8 & 22.1 & 24.3 & 33.0 & 26.0 & 24.1 & 32.1 \\
     & & HNN (Avg) &  \multicolumn{7}{c|}{ \rule[0.25em]{0.3\columnwidth}{0.3pt} 39.3 \rule[0.25em]{0.3\columnwidth}{0.3pt} } & 39.6 & 37.9 & 33.6 & 42.2 & 39.3 & 34.6 & 43.1 \\
      &  & HAN (Att)  &  \multicolumn{7}{c|}{ \rule[0.25em]{0.3\columnwidth}{0.3pt} 43.4 \rule[0.25em]{0.3\columnwidth}{0.3pt} } & 44.8 & 46.3 & 41.9 & 46.4 & 45.8 & 41.2 & 49.4 \\ \cline{2-17}
    &  \parbox[t]{2mm}{\multirow{3}{*}{\rotatebox[origin=c]{90}{  Multi  }}}  & MHAN-Enc  & 45.4 & 45.9 & 44.3 & 41.1 & 42.1 & 44.9 & 41.0 & 43.9 & 46.2 & 39.3 & 47.4 & 45.0 & 37.9 & 48.6  \\
  & & MHAN-Att &  \textbf{46.3} & \textbf{46.0} & \textbf{45.9} & \textbf{45.6} & \textbf{46.4} & \textbf{46.4} & \textbf{46.1} & \textbf{46.5} & \textbf{46.7} & \textbf{43.3} & \textbf{47.9} & {45.8} & \textbf{41.3} & 48.0 \\
  & &  MHAN-Both &  45.7 & 45.6 & 41.5 & 41.2 & 45.6 & 44.6 & 43.0 & 45.9 & 46.4 & 40.3 & 46.3 & \textbf{46.1} & 40.7 & \textbf{50.3} \\
	\hline
\end{tabular}
\caption{\label{full-rec_results}Full-resource classification performance ($F_{1}$) on general (top) and specific (bottom) topic categories using bilingual training with English as target (left) and the auxiliary language as target (right).
}
\end{table*}

\subsection{Training over Disjoint Label Sets}
\label{multi_strategy}

For training, we replace the monolingual training objective (Eq.~\ref{eq:mono-objective}) with a joint multilingual objective that facilitates the sharing of components, i.e.\ a subset of parameters for each language $\theta_1, \ldots, \theta_M$, across different language networks:
\begin{equation}
\vspace{-0.5em}
  \label{multilingual_objective}
\begin{split}
  \mathcal{L}(\theta_1, \ldots, \theta_{M} )
 = - \frac{1}{Z}  \sum^{N_e}_{i} \sum^{M}_{l} \mathcal{H}(y^{(l)}_{i}, \hat{y}^{(l)}_{i})
\end{split}
\end{equation}
\noindent where $Z = M \times N_e$ and $N_e$ is the epoch size.%
\footnote{In the future, it may also be beneficial to add a $\gamma_{l}$ term for each language objective, which encodes prior knowledge about its importance.}

The joint objective $\mathcal{L}$ can be minimized with respect to the parameters $\theta_1, \ldots, \theta_M$ using SGD as before. However, when training on examples from different languages consecutively it is difficult to learn a shared space that works well across languages. This is because updates for each language apply only on a subset of parameters and may bias the model away from other languages. To address this issue, we employ the training strategy proposed by \citep{firat16}, who sampled parallel sentences for multi-way machine translation from different language pairs in a cyclic fashion at each iteration.\footnote{We  verified this empirically in our preliminary experiments and found that mixing languages in a single batch performed better than keeping them in separate batches.}  Here, we sample a document-label pair from each language at iteration.  For minibatch SGD, the number of samples per language is equal to the batch size divided by M.

\section{A New Corpus for Multilingual Document Classification: DW}
\label{sec:dw-dataset}

Multilingual document classification datasets are usually limited in size, have target categories aligned across languages, and assign documents to only one category. However, classification is often necessary in cases where the categories are not strictly aligned, and multiple categories may apply to each document.
 For instance, this is the case for online multilingual news agencies, which must keep track of news topics across languages.

Two datasets for multilingual document classification have been used in previous studies: Reuters RCV1/RCV2 (6,000 documents, 2 languages and 4 labels), introduced by \cite{klementiev12}, and TED talk transcripts (12,078 documents, 12 languages and 15 labels), introduced by \citet{hermann14}. The former is tailored for evaluating word embeddings aligned across languages, rather than complex multilingual document models.  The latter is twice as large and covers more languages, in a multi-label setting, but biases evaluation by including translations of talks in all languages.

Here, we present and use a much larger dataset collected from Deutsche Welle, Germany's public international broadcaster, shown in Table~\ref{dw_stats}.
The DW dataset contains nearly 600,000 documents, in 8 languages, annotated by journalists with several topic labels. Documents are on average 2.6 times longer than in Yang et al.'s (\citeyear{yang16}) monolingual dataset (436 vs.\ 163 words).
There are two types of labels, namely \textit{general topics} ($Y_g$) and \textit{specific} ones ($Y_s$) both described by one or more words. We consider (and count in Table~\ref{dw_stats}) only those specific labels that appear at least 100 times, to avoid sparsity issues. 

The number of labels varies greatly across the 8 languages. Moreover, we found for instance that only 25-30\% of the labels could be manually aligned between English and German.  The commonalities are mainly concentrated on the most frequent labels, reflecting a shared top-level division of the news domain, but the long tail exhibits significant independence across languages.

\section{Evaluation}
\label{sec:eval}

\subsection{Settings}
\label{sec:exp-settings}
We evaluate our multilingual models on full-resource and low-resource scenarios of multilingual document classification on the Deutsche Welle corpus. Following the typical evaluation protocol in the field, the corpus is split per language into 80\% for training, 10\% for validation and 10\% for testing. We evaluate both type of labels ($Y_{g}$, $Y_{s}$) on a \textit{full-resource scenario} and only the general topic labels ($Y_{g}$) on a \textit{low-resource scenario}.  We report the micro-averaged F1 scores  for each test set, as in previous work \citep[e.g.,][]{hermann14}.

\textbf{Model configuration.} For all models, we use the aligned 40-dimensional multilingual embeddings pre-trained on the Leipzig corpus using multi-CCA from \citet{ammar16}. The non-aligned embeddings used for comparison purposes are trained with the same method and data.  We zero-pad documents up to a maximum of 30 words per sentence and 30 sentences per document. The hyper-parameters were selected on the validation sets. We made the following settings: 100-dimensional encoder and attention embeddings (at every level), $\mathrm{relu}$ activation function for all intermediate layers, batch size of 16, epoch size of 25k, and optimization using SGD with Adam until convergence.

\begin{figure*}
 \centering
 \includegraphics[scale=0.392]{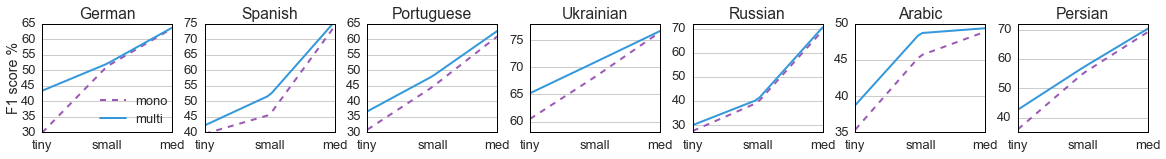}\
 \vspace{-7mm}
 \caption{\label{low-rec_summary} Low-resource document classification performance ($F_1$) of our \textit{multilingual} attention network ensemble (blue lines) vs.\ a \textit{monolingual} attention network (purple dashed lines) on the DW corpus.}
 \vspace{-2mm}
 \end{figure*}

All the hierarchical models have \textsc{Dense} encoders in both scenarios (Tables~\ref{full-rec_results}, \ref{multi8}, and~\ref{low-rec_results}), and GRU and biGRU in the full-resource scenario for English+Arabic (Table~\ref{gru}). For the low-resource scenario, we define three levels of  data availability:  \textit{tiny} from 0.1\% to 0.5\%, \textit{small} from 1\% to 5\% and \textit{medium} from 10\% to 50\% of the original training set.
 We report the average $F_{1}$ scores on the test set for each level based on discrete increments of 0.1, 1 and 10 percentage points respectively.  The decision threshold for the value of $p$ in Eq.~\ref{eq:classification} for the full-resource scenario is set to 0.4 for labels such that  $|Y_{s}|<400$ and 0.2 for $|Y_{s}|\geq400$, and for the low-resource scenario it is 0.3 for all sets. For the \emph{ensemble} in the low-resource setting, we train the three proposed multilingual models and choose the optimal one based on the validation data for each language respectively (see Fig.~\ref{low-rec_summary}).

\textbf{Baselines.} We compare against the following monolingual neural networks, with shallow or hierarchical structures.  These networks are based on the state of the art in the field, reviewed in Section~\ref{sec:related-work}, and thus represent strong baselines.
\begin{itemize} \setlength{\itemsep}{-0.05em}
  \item \textbf{NN }: A neural network which feeds the average vector of the input words directly to a classification layer, as the common baseline for multilingual document classification \citep{klementiev12}.
  \item \textbf{HNN }: A hierarchical network with encoders and average pooling at every level,	followed by a classification layer.
  \item  \textbf{HAN}: A hierarchical network with encoders and attention, followed by a classification layer. This model is the one proposed by \citet{yang16} adapted to our task.
\end{itemize}
Our multilingual models with the three sharing configurations from Section~\ref{sec:sharing-components}, are noted as \textit{Enc}, \textit{Att} and \textit{Both}.  Their implementation amounts to, first, creating a HAN model for each language, second, sharing components across multiple languages as illustrated in Fig.~\ref{multilingual_architecture}, and, third, training them with the objective of Eq.~\ref{multilingual_objective}.


\subsection{Results}
\label{sec:results}

\textbf{Full-resource scenario.} Table~\ref{full-rec_results} displays the results of full-resource document classification using \textsc{Dense} encoders for general and specific labels.  On the left side, the performance on the English sub-corpus is shown when English and an auxiliary sub-corpus  are used for training, and on the right side, the performance on the auxiliary sub-corpus is shown when that sub-corpus and the English sub-corpus are used for training.

The multilingual model trained on pairs of languages outperforms on average all the examined monolingual models, namely a bag-of-word neural model and two  hierarchical neural models which use average pooling and attention respectively.
The best-performing multilingual model bilingually on average is the one with shared attention across languages, especially when tested on English. The consistent gain for English as target could be attributed to the alignment of the word embeddings to English and to the many English labels, which makes it easier to find multilingual labels from which to transfer knowledge. Interestingly, this reveals that the transfer of knowledge across languages in a full-resource setting is maximized with language-specific word and sentence encoders, but language-independent (i.e.\ shared) attention for both words and sentences.

However, when transferring from English to Portuguese (en$\rightarrow$pt), Russian (en$\rightarrow$ru) and Persian (en$\rightarrow$fa) on general categories, it is more effective to have only language-independent components.  We hypothesize that this is due to the underlying commonness between the label sets rather than to a relationship between languages, which is hard to identify on linguistic grounds.

 \begin{table}[ht]
 \small
 \centering
   \begin{tabular}{ c | l | c | c  c c c}
     &  Encoders &  \multicolumn{1}{c|}{Mono}  & \multicolumn{3}{c}{Multi}  \\ \hline
   & $Y_\mathit{general}$ & HAN & Enc & Att & Both  \\
     \hline
      \parbox[t]{2mm}{\multirow{3}{*}{\rotatebox[origin=c]{90}{{ar}$\rightarrow${en}}}} & \textsc{Dense} & 71.2 & 70.0 & \textbf{73.8} & 68.9\\
      & GRU & 77.0 & 74.8 & \textbf{77.5} & 75.4   \\
      & biGRU &  \textbf{~77.7} & 77.1 & {77.5} & 76.7\\ \hline
      \parbox[t]{2mm}{\multirow{3}{*}{\rotatebox[origin=c]{90}{{en}$\rightarrow${ar}}}} & \textsc{Dense} &  80.5 & 79.0 & \textbf{82.1} & 79.1\\
      & GRU & 81.5  & 81.2 & \textbf{83.4} & 83.1 \\
      & biGRU & 82.2 & 82.7 & ~\textbf{84.0}  & 83.0 \\
      \hline
 \end{tabular}
 \caption{\label{gru} Full-resource classification performance ($F_{1}$) for English-Arabic with various encoders.}
 \vspace{-3mm}
 \end{table}
 \begin{table}[ht]
  \def\arraystretch{1.3}
  \setlength{\tabcolsep}{2.5pt}
  \small
  \centering
 \begin{tabular}{ c | c | c | c | c | c }
   \multicolumn{2}{c|}{ }  & \multicolumn{2}{c|}{$Y_\mathit{general}$}   & \multicolumn{2}{c}{$Y_\mathit{specific}$}  \\ \hline
      Word embeddings & $|L|$  &   $ n_{l}$ & $f_{l}$  &   $ n_{l}$ & $f_{l}$  \\ \hline
         &  1 & 50K --  & 77.41 -- & 90K -- & 44.90 --\\
    Aligned    &  2 & 40K $\downarrow$ & 78.30 $\uparrow$ & 80K $\downarrow$ & 45.72 $\uparrow$ \\
        &  8 & 32K $\downarrow$ & 77.91 $\uparrow$  & 72K$\downarrow$ & 45.82 $\uparrow$ \\ \hline
  Non-aligned      &  8 & 32K $\downarrow$ & 71.23 \textcolor{red}{$\downarrow$}  & 72K $\downarrow$ & 33.41 \textcolor{red}{$\downarrow$} \\
  \hline
 \end{tabular}
 \caption{\label{multi8}Average number of parameters per language ($n_{l}$), average $F_1$ per  language ($f_{l}$), and their variation (arrows) with the number of languages $|L|$ and the word embeddings used for training.}
 \vspace{-1mm}
 \end{table}
We will now quantify the impact of three important model choices on the performance: encoder type, word embeddings, and number of languages used for training. In Table~\ref{gru}, we observe that when we replace the \textsc{Dense} encoder layers  with GRU or biGRU layers, the improvement from the multilingual training is still present. In particular, the multilingual models with shared attention are superior to alternatives, regardless of the employed encoders.  For reference, using simply logistic regression with bag-of-words (counts) for classification leads to $F_1$ scores of 75.8\% in English and 81.9\% in Arabic, using many more parameters than biGRU: 56.5M vs.\ 410k in English and 5.8M vs.\ 364k in Arabic.

In Table~\ref{multi8},  when we train our multilingual model (MHAN-att) on eight languages at the same time, the $F_{1}$ score improves on average across languages -- for both types of labels, general or specific -- while the number of parameters per language decrease, by 36\% for $Y_\mathit{general}$ and 20\% for $Y_\mathit{specific}$. Lastly, when we train the same model with word embeddings that are not aligned across languages, the performance of the multilingual model drops significantly.  An input space that is aligned across languages is thus crucial.

\textbf{Low-resource scenario.} We assess the ability of the multilingual attention networks to transfer knowledge across languages in a low-resource scenario, i.e.\ training on a fraction of the available data, as defined in~\ref{sec:exp-settings} above.  The results for seven languages when trained jointly with English are displayed in detail in Table~\ref{low-rec_results} and summarized in Figure~\ref{low-rec_summary}. In all cases, at least one of the multilingual models outperforms the monolingual one, which demonstrates the usefulness of multilingual training for low-resource document classification.

\begin{table}[t]
  \setlength{\tabcolsep}{4.5pt}
	\small
\centering
  \begin{tabular}{ c | c | c | c  c c | l}
   & Size  &  \multicolumn{1}{c|}{Mono}  & \multicolumn{3}{c}{Multi}  \\ \hline
      & $Y_\mathit{general}$  &  HAN & Enc & Att & Both & $\Delta$\% \\
    \hline
     \parbox[t]{3mm}{\multirow{3}{*}{\rotatebox[origin=c]{90}{{en}$\rightarrow${de}}}} & 0.1-0.5\% & 29.9 & \textbf{41.0} & 37.0  & 39.4 & +37.2\\
     & 1-5\% & 51.3 & 51.7 & 49.7 & \textbf{52.6}  &+2.6\\
     & 10-50\% & 63.5 & 63.0 & {63.8} & \textbf{63.8}  &+0.5 \\ \hline

     \parbox[t]{2mm}{\multirow{3}{*}{\rotatebox[origin=c]{90}{{en}$\rightarrow${es}}}} & 0.1-0.5\% & 39.5 & 38.7 & 33.3  & \textbf{41.5} & +4.9 \\
      & 1-5\%     & 45.6 & 45.5 & \textbf{50.8} & 50.1 & +11.6 \\
      & 10-50\%   & 74.2 & \textbf{75.7} & 74.2 & 75.2 & +2.0 \\ \hline

     \parbox[t]{2mm}{\multirow{3}{*}{\rotatebox[origin=c]{90}{{en}$\rightarrow${pt}}}} & 0.1-0.5\% & 30.9 & 25.3 &  31.6 & \textbf{33.8} & +9.6 \\
     & 1-5\% & 44.6 & 44.3 & 37.5 & \textbf{47.3} & +6.0  \\
     & 10-50\% & 60.9 & 61.9 & 62.1 & \textbf{62.1} & +1.9\\ \hline

     \parbox[t]{3mm}{\multirow{3}{*}{\rotatebox[origin=c]{90}{{en}$\rightarrow${uk}}}} & 0.1-0.5\% & 60.4 & \textbf{62.4} & 59.8 & 60.9 & +3.1\\
      & 1-5\%     & 68.2 & 67.7 & \textbf{70.6} & 69.0 & +3.4\\
      & 10-50\%   & 76.4 & 76.2 & 76.3 & \textbf{76.7} & +0.3\\ \hline

     \parbox[t]{2mm}{\multirow{3}{*}{\rotatebox[origin=c]{90}{{en}$\rightarrow${ru}}}} & 0.1-0.5\% & 27.6 & 26.6 & 27.0 & \textbf{29.1} & +5.4 \\
      & 1-5\%     & 39.3 & 38.2 & 39.6 & \textbf{40.2} & +2.2 \\
      & 10-50\%   & 69.2 & \textbf{70.5} & 70.4 & {69.4} & +1.9 \\ \hline

     \parbox[t]{2mm}{\multirow{3}{*}{\rotatebox[origin=c]{90}{{en}$\rightarrow${ar}}}} & 0.1-0.5\% & 35.4 & 35.5 & \textbf{39.5} & 36.6 & +11.7\\
     & 1-5\%      & 45.6 & \textbf{48.7} & 47.2 & 46.6 & +6.9\\
     & 10-50\%    & 48.9 & \textbf{52.2} & 46.8 & 47.8 & +6.8\\ \hline

     \parbox[t]{3mm}{\multirow{3}{*}{\rotatebox[origin=c]{90}{{en}$\rightarrow${fa}}}} & 0.1-0.5\% & 36.0 & 35.6 & 33.6 & \textbf{41.3} & +14.6\\
     & 1-5\%      & 55.0 & \textbf{55.6} & 51.9 & 55.5 & +1.0\\
     & 10-50\%    & 69.2 & \textbf{70.3} & 70.1 & 70.0 & +1.5 \\ \hline
  \end{tabular}
\caption{\label{low-rec_results}Low-resource classification performance ($F_{1}$) with various sizes of training data.}
\vspace{-5mm}
\end{table}

Moreover, the improvements obtained from our multilingual models for lower levels of availability (\textit{tiny} and \textit{small}) are
larger than in higher levels (\textit{medium}).  This is also clearly observed in Figure~\ref{low-rec_summary} with our multilingual attention network ensemble, i.e.\ when we do model selection among the three multilingual variants on the development set. The best performing architecture in a majority of cases is the one which shares both the encoders and the attention mechanisms across languages.  Moreover, this architecture also has the fewest number of parameters.

This promising finding for the low-resource scenario means that the classification performance can greatly benefit from the multilingual training (sharing encoders and attention) without increasing the number of parameters beyond that of a single monolingual document model. Nevertheless, in a few cases, we observe that the other architectures with increased complexity perform better than the ``shared both'' model.
For instance, sharing encoders is superior to alternatives for Arabic language, i.e.\ the knowledge transfer benefits from shared word and sentence  representations. Hence, to generalize to a large number of languages, we may need to consider more dynamic models which are able to choose for each language individually which sharing scheme is the most appropriate for transferring from  another language.  Lastly, we did not generally observe a negative (or positive) correlation of the similarity between languages with the performance in the low-resource scenario, although the largest improvements were observed on languages more related to English (German, Spanish, Portuguese) than others (Arabic).

Overall, the above experiments pinpointed the most suitable multilingual sharing scheme (Figure~\ref{multilingual_architecture}) for each setting independently of the encoder type, rather than the optimal combination of sharing scheme and encoder. Therefore, as shown in Table~\ref{gru}, increasing the sophistication of the encoders (from \textsc{Dense} to GRU to biGRU) is expected to further improve accuracy.

\subsection{Qualitative Analysis}

\begin{figure}[t]
\centering
\includegraphics[scale=0.17]{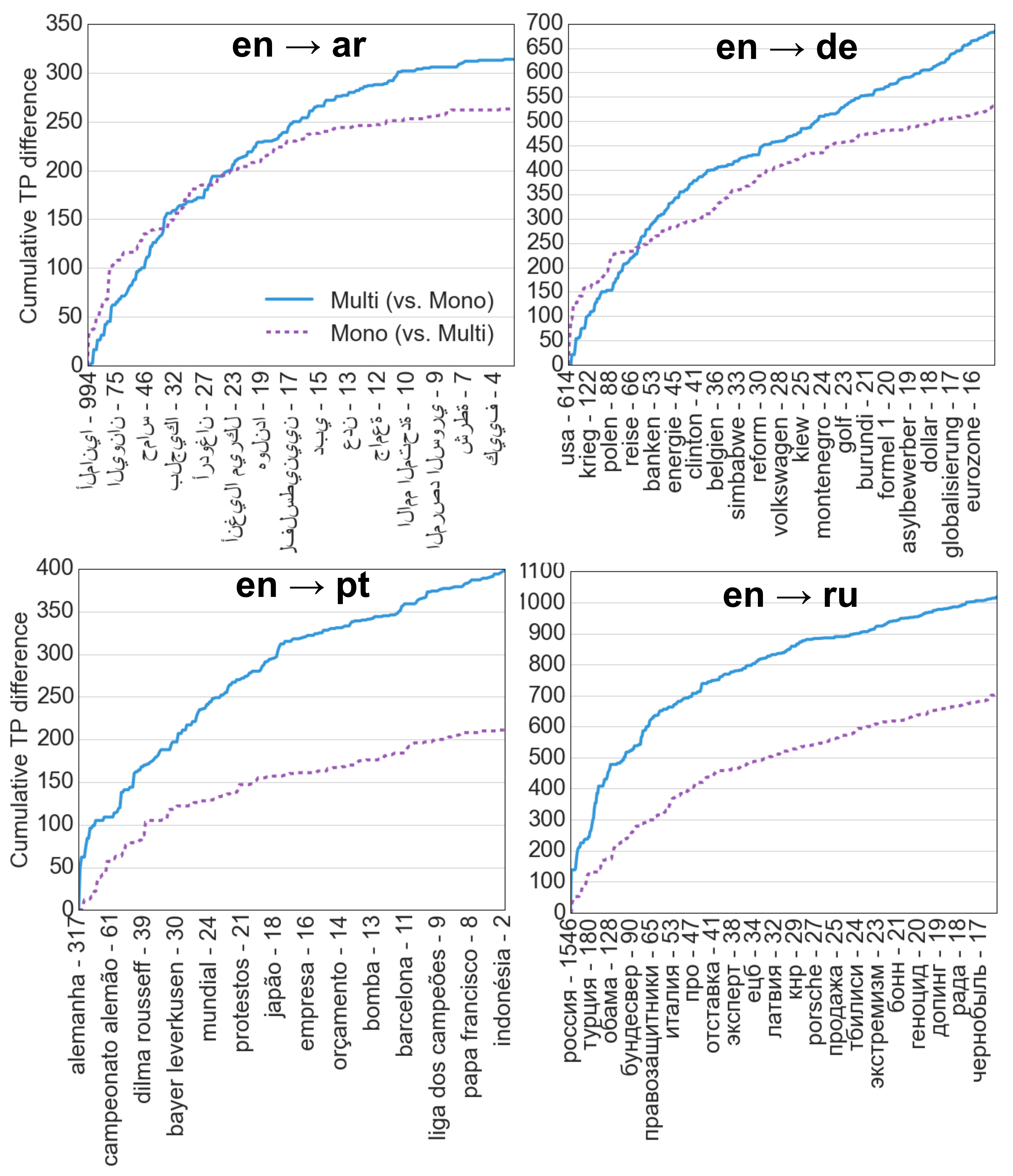}
\vspace{-6mm}
\caption{\label{cumulative_tp}Cumulative true positive (TP) difference between \textit{monolingual} and \textit{multilingual} (ensemble) models for topic classification with \textit{specific} labels, in the full resource scenario.}
\vspace{-2mm}
\end{figure}

We analyze the performance of the multilingual model over the full range of labels, to observe on which type of labels it performs better than the monolingual model, and provide some qualitative examples. Figure~\ref{cumulative_tp} shows the cumulative true positive (TP) difference between the monolingual and multilingual models on the Arabic, German, Portuguese and Russian test sets, ordered by label frequency.  We can observe that the cumulative TP difference of the multilingual model consistently increases  as the frequencies of the labels decrease. This shows that labels across the entire range of frequencies benefit from joint training with English and not only a subset, for example only the highly frequent labels.

For example, the top 5 labels on which the multilingual model performed better than the monolingual one for en$\rightarrow$de were: \textit{russland} (21), \textit{berlin} (19), \textit{irak} (14), \textit{wahlen} (13) and \textit{nato} (13), while for the opposite direction those were: \textit{germany} (259), \textit{german} (97), \textit{soccer} (73), \textit{football} (47) and \textit{merkel} (25).  These topics are likely better covered in the respective auxiliary language which helps the multilingual model to better distinguish them in the target language as well. This is also observed in Figure~\ref{tsne_projection} presented in the introduction, through an improved separation of topics using multilingual model vs.\ monolingual ones.

\section{Conclusion}

We proposed multilingual hierarchical attention networks for document classification and showed that they can benefit both full-resource and low-resource  settings, while using fewer parameters than monolingual networks.  In the former setting, the best option was to share only the attention mechanisms, while in the latter one, it was sharing the encoders along with the attention mechanisms.  These results confirm the merits of language transfer, which is also an important component of human language learning \citep{Odlin_1989,ringbom2007}. Moreover, our study broadens the applicability of multilingual document classification, since our framework is not restricted to common label sets.

There are several future directions for this study.  In their current form, our models cannot generalize to languages without any example, as attempted by \citet{firat16b} for neural machine translation. This could be achieved by a classification layer independent of the size of the label set as in zero-shot classification \citep{qiao16,nam16}. Moreover, although we explored three distinct architectures, other configurations could be examined to improve document modeling, for example by sharing the attention mechanism at the sentence-level only. Lastly, the learning objective could be further constrained with sentence-level parallel information, to embed multilingual vectors of similar topics more closely together in the learned space.

\section*{Acknowledgments}

We are grateful for the support from the European Union through its Horizon 2020 program in the SUMMA project n.\ 688139, see \url{http://www.summa-project.eu}.  We would also like to thank Sebasti\~{a}o Miranda at Priberam for gathering the news articles from Deutsche Welle and the anonymous reviewers for their helpful suggestions.  The second author contributed to the paper while at the Idiap Research Institute.

\bibliography{database}
\bibliographystyle{ijcnlp2017}
\end{document}